\documentclass[10pt,twocolumn,letterpaper]{article}

\usepackage{algorithm}
\usepackage{algorithmic}
\usepackage{bm}
\usepackage{amsthm}
\usepackage{multirow, paralist}
\usepackage{color}
\usepackage{bbm}

\def \R {\mathbb{R}}

\def \x {\mathbf{x}}
\def \OO {\mathcal{O}}
\def \G {\mathcal{G}}
\def \F {\mathcal{F}}
\def \H {\mathcal{H}}
\def \G {\mathcal{G}}

\def \z {\mathbf{z}}

\def \D {\mathcal{D}}

\def \LL {\mathcal{L}}
\def \S {\mathcal{S}}

\newtheorem{thm}{Theorem}

\usepackage{cvpr}
\usepackage{times}
\usepackage{epsfig}
\usepackage{graphicx}
\usepackage{amsmath}
\usepackage{amssymb}

\usepackage[pagebackref=true,breaklinks=true,letterpaper=true,colorlinks,bookmarks=false]{hyperref}

\cvprfinalcopy 

\def\cvprPaperID{3381} 

\ifcvprfinal\fi
\begin{document}

\title{Large-scale Distance Metric Learning with Uncertainty}

\author{Qi Qian\quad\quad   Jiasheng Tang\quad\quad  Hao Li\quad\quad    Shenghuo Zhu\quad\quad    Rong Jin\\
Alibaba Group, Bellevue, WA, 98004, USA\\
{\tt\small \{qi.qian, jiasheng.tjs, lihao.lh, shenghuo.zhu, jinrong.jr\}@alibaba-inc.com}
}

\maketitle
\thispagestyle{empty}

\begin{abstract}
Distance metric learning (DML) has been studied extensively in the past decades for its superior performance with distance-based algorithms. Most of the existing methods propose to learn a distance metric with pairwise or triplet constraints. However, the number of constraints is quadratic or even cubic in the number of the original examples, which makes it challenging for DML to handle the large-scale data set. Besides, the real-world data may contain various uncertainty, especially for the image data. The uncertainty can mislead the learning procedure and cause the performance degradation. By investigating the image data, we find that the original data can be observed from a small set of clean latent examples with different distortions. In this work, we propose the margin preserving metric learning framework to learn the distance metric and latent examples simultaneously. By leveraging the ideal properties of latent examples, the training efficiency can be improved significantly while the learned metric also becomes robust to the uncertainty in the original data. Furthermore, we can show that the metric is learned from latent examples only, but it can preserve the large margin property even for the original data. The empirical study on the benchmark image data sets demonstrates the efficacy and efficiency of the proposed method.
\end{abstract}

\section{Introduction}
\label{sec:intro}

\begin{figure}[!ht]\label{fig:illu}
\centering
\includegraphics[width=3in]{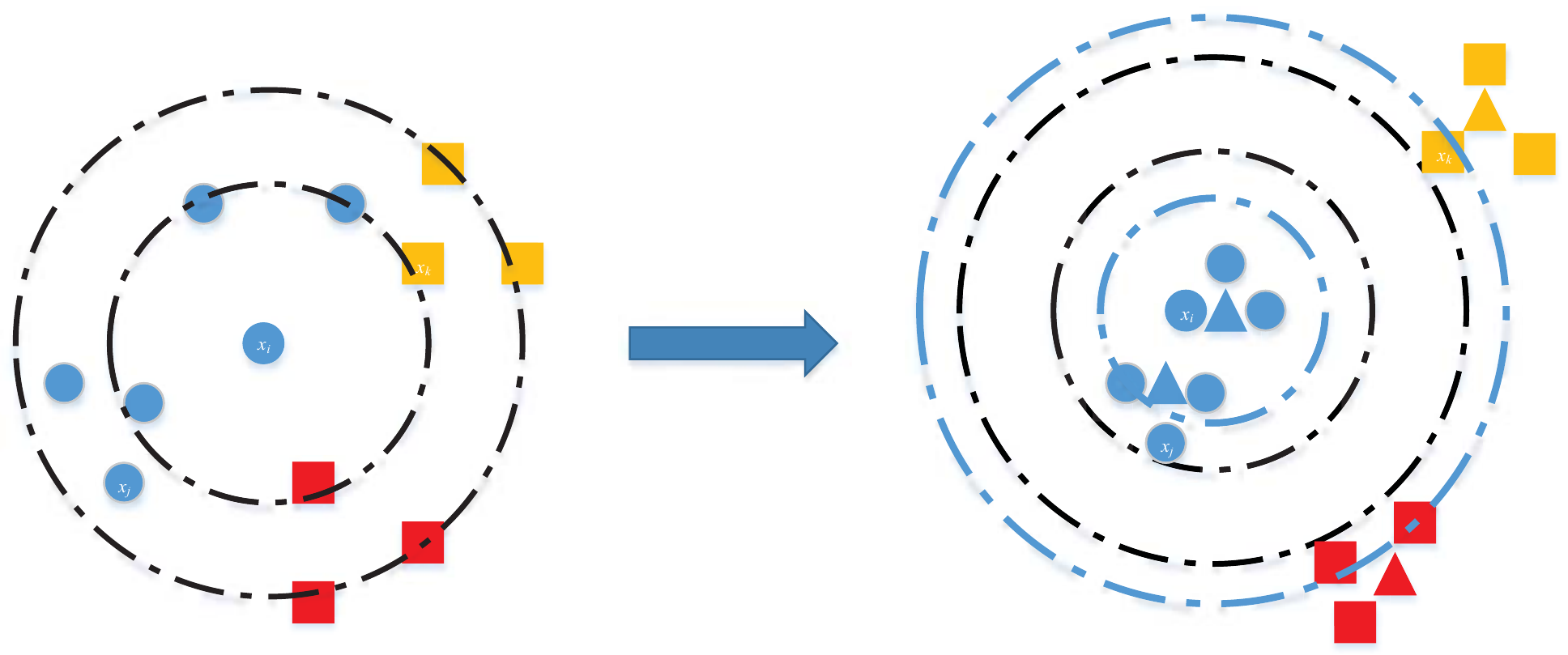}
\caption{Illustration of the proposed method. Let round and square points denote the target data and impostors, respectively. Let triangle points denote the corresponding latent examples. Data points with the same color are from the same class. It demonstrates that the metric learned with latent examples not only separates the dissimilar latent data with a large margin but also preserves the large margin for the original data.}
\end{figure}

Distance metric learning (DML) aims to learn a distance metric where examples from the same class are well separated from examples of different classes. It is an essential task for distance-based algorithms, such as $k$-means clustering~\cite{XingNJR02}, $k$-nearest neighbor classification~\cite{weinberger2009} and information retrieval~\cite{chechik2010}. Given a distance metric $M$, the squared Mahalanobis distance between examples $\x_i$ and $\x_j$ can be computed as
\[\D_M^2(\x_i,\x_j) = (\x_i-\x_j)^\top M(\x_i-\x_j)\]
Most of existing DML methods propose to learn the metric by minimizing the number of violations in the set of pairwise or triplet constraints. Given a set of pairwise constraints, DML tries to learn a metric such that the distances between examples from the same class are sufficiently small (e.g., smaller than a predefined threshold) while those between different ones are large enough~\cite{DavisKJSD07,XingNJR02}. Different from pairwise constraints, each triplet constraint consists of three examples $(\x_i,\x_j,\x_k)$, where $\x_i$ and $\x_j$ have the same label and $\x_k$ is from a different class. An ideal metric can push away $\x_k$ from $\x_i$ and $\x_j$ by a large margin~\cite{weinberger2009}. Learning with triplet constraints optimizes the local positions of examples and is more flexible for real-world applications, where defining the appropriate thresholds is hard for pairwise constraints. In this work, we will focus on DML with triplet constraints.

Optimizing the metric with a set of triplet constraints is challenging since the number of triplet constraints can be up to $\OO(n^3)$, where $n$ is the number of the original training examples. It makes DML computationally intractable for the large-scale problems. Many strategies have been developed to deal with this challenge and most of them fall into two categories, learning by stochastic gradient descent (SGD) and learning with the active set. With the strategy of SGD, DML methods can sample just one constraint or a mini-batch of constraints at each iteration to observe an unbiased estimation of the full gradient and avoid computing the gradient from the whole set~\cite{chechik2010,QianJY0Z15}. Other methods learn the metric with a set of active constraints (i.e., violated by the current metric), where the size can be significantly smaller than the original set~\cite{weinberger2009}. It is a conventional strategy applied by cutting plane methods~\cite{Boyd}. Both of these strategies can alleviate the large-scale challenge but have inherent drawbacks. Approaches based on SGD have to search through the whole set of triplet constraints, which results in the slow convergence, especially when the number of active constraints is small. On the other hand, the methods relying on the active set have to identify the set at each iteration. Unfortunately, this operation requires computing pairwise distances with the current metric, where the cost is $\OO(n^2)$ and is too expensive for large-scale problems. 

Besides the challenge from the size of data set, the uncertainty in the data is also an issue, especially for the image data, where the uncertainty can come from the differences between individual examples and distortions, e.g., pose, illumination and noise. Directly learning with the original data will lead to a poor generalization performance since the metric tends to overfit the uncertainty in the data. By further investigating the image data, we find that most of original images can be observed from a much smaller set of clean latent examples with different distortions. The phenomenon is illustrated in Fig.~\ref{fig:mexample}. This observation inspires us to learn the metric with latent examples in lieu of the original data. The challenge is that latent examples are unknown and only images with uncertainties are available.

In this work, we propose a framework to learn the distance metric and latent examples simultaneously. It sufficiently explores the properties of latent examples to address the mentioned challenges. First, due to the small size of latent examples, the strategy of identifying the active set becomes affordable when learning the metric. We adopt it to accelerate the learning procedure via avoiding the attempts on inactive constraints. Additionally, compared with the original data, the uncertainty in latent examples decreases significantly. Consequently, the metric directly learned from latent examples can focus on the nature of the data rather than the uncertainty in the data. To further improve the robustness, we adopt the large margin property that latent examples from different classes should be pushed away with a data dependent margin. Fig.~\ref{fig:illu} illustrates that an appropriate margin for latent examples can also preserve the large margin for the original data. We conduct the empirical study on benchmark image data sets, including the challenging ImageNet data set, to demonstrate the efficacy and efficiency of the proposed method.

The rest of the paper is organized as follows: Section \ref{sec:related} summarizes the related work of DML. Section \ref{sec:method} describes the details of the proposed method and Section \ref{sec:analysis} summarizes the theoretical analysis. Section \ref{sec:exp} compares the proposed method to the conventional DML methods on the benchmark image data sets. Finally, Section \ref{sec:conclusion} concludes this work with future directions.

\section{Related Work}
\label{sec:related}

Many DML methods have been proposed in the past decades~\cite{DavisKJSD07,weinberger2009,XingNJR02} and comprehensive surveys can be found in \cite{Kulis13,liu2006}. The representative methods include Xing's method~\cite{XingNJR02}, ITML~\cite{DavisKJSD07} and LMNN~\cite{weinberger2009}. ITML learns a metric according to pairwise constraints, where the distances between pairs from the same class should be smaller than a predefined threshold and the distances between pairs from different classes should be larger than another predefined threshold. LMNN is developed with triplet constraints and a metric is learned to make sure that pairs from the same class are separated from the examples of different classes with a large margin. Compared with pairwise constraints, triplet constraints are more flexible to depict the local geometry.

To handle the large number of constraints, some methods adopt SGD or online learning to sample one constraint or a mini-batch of constraints at each iteration~\cite{chechik2010,QianJY0Z15}. OASIS~\cite{chechik2010} randomly samples one triplet constraint at each iteration and computes the unbiased gradient accordingly. When the size of the active set is small, these methods require extremely large number of iterations to improve the model. Other methods try to explore the concept of the active set. LMNN~\cite{weinberger2009} proposes to learn the metric effectively at each iteration by collecting an active set that consists of constraints violated by the current metric within the $k$-nearest neighbors for each example. However, it requires $\OO(n^2)$ to obtain the appropriate active set. 

Besides the research about conventional DML, deep metric learning has attracted much attention recently~\cite{Attias17,RippelPDB15,SchroffKP15,SongXJS16}. These studies also indicate that sampling active triplets is essential for accelerating the convergence. FaceNet~\cite{SchroffKP15} keeps a large size of mini-batch and searches hard constraints within a mini-batch. LeftedStruct~\cite{SongXJS16} generates the mini-batch with the randomly selected positive examples and the corresponding hard negative examples. Proxy-NCA~\cite{Attias17} adopts proxy examples to reduce the size of triplet constraints. Once an anchor example is given, the similar and dissimilar examples will be searched within the set of proxies. In this work we propose to learn the metric only with latent examples which can dramatically reduce the computational cost of obtaining the active set. Besides, the triangle inequality dose not hold for the squared distance, which makes our analysis significantly different from the existing work.

\section{Margin Preserving Metric Learning}
\label{sec:method}
Given a training set $\{(\x_i,y_i)|i=1,\cdots,n\}$, where $\x_i\in\R^d$ is an example and $y_i$ is the corresponding label, DML aims to learn a good distance metric such that
\[\forall \x_i,\x_j,\x_k\quad \D_M^2(\x_i,\x_k)-\D_M^2(\x_i,\x_j)\geq 1\] 
where $\x_i$ and $\x_j$ are from the same class and $\x_k$ is different.
Given the distance metric $M\in \S_+^{d\times d}$, the squared distance is defined as
\[\D_M^2(\x_i,\x_j) = (\x_i-\x_j)^\top M(\x_i-\x_j)\]
where $\S_+^{d\times d}$ denotes the set of $d\times d$ positive semi-definite (PSD) matrices.

For the large-scale image data set, we assume that each observed example is from a latent example with certain zero mean distortions, i.e.,
\[\forall i,\quad E[\x_i] = \z_{o:f(i)=o}\]
where $f(\cdot)$ projects the original data to its corresponding latent example.

Then, we consider the expected distance~\cite{YeZSJ17} between observed data and the objective is to learn a metric such that
\begin{eqnarray}\label{eq:ocons}
\forall \x_i,\x_j,\x_k\quad E[\D_M^2(\x_i,\x_k)]-E[\D_M^2(\x_i,\x_j)]\geq 1
\end{eqnarray}

Let $\z_o$, $\z_p$ and $\z_q$ denote latent examples of $\x_i$, $\x_j$ and $\x_k$ respectively.
For the distance between examples from the same class, we have
\begin{align}\label{eq:sim}
&E[\D_M^2(\x_i,\x_j)] = E[(\x_i-\z_o+\z_o)^\top M(\x_i-\z_o+\z_o)]\nonumber\\
&+E[(\x_j-\z_p+\z_p)^\top M (\x_j-\z_p+\z_p)] - E[2\x_i^\top M \x_j]\nonumber\\
& = \D_M^2(\z_o,\z_p)+E [\D_M^2(\x_i,\z_o)]+E[\D_M^2(\x_j,\z_p)]\nonumber\\
& = \D_M^2(\z_o,\z_p)+2E [\D_M^2(\x_i,\z_o)]
\end{align}
The last equation is due to the fact that $\x_i$ and $\x_j$ are i.i.d, since they are from the same class.

By applying the same analysis for the dissimilar pair, we have
\begin{align}\label{eq:dis}
&E[\D_M^2(\x_i,\x_k)] = \D_M^2(\z_o,\z_q)+E [\D_M^2(\x_i,\z_o)]\nonumber\\
&+E [\D_M^2(\x_k,\z_q)]\geq  \D_M^2(\z_o,\z_q)+E [\D_M^2(\x_i,\z_o)]
\end{align}
The inequality is because that $M$ is a PSD matrix.

Combining Eqns.~\ref{eq:sim} and \ref{eq:dis}, we find that the difference between the distances in the original triplet can be lower bounded by those in the triplet consisting of latent examples
\begin{eqnarray*}
&&E[\D_M^2(\x_i,\x_k)]-E[\D_M^2(\x_i,\x_j)]\\
&&\geq \D_M^2(\z_o,\z_q)-\D_M^2(\z_o,\z_p) - E [\D_M^2(\x_i,\z_o)]
\end{eqnarray*}

Therefore, the metric can be learned with the constraints defined on latent examples such that
\[\forall \z_o,\z_p,\z_q\ \ \ \D_M^2(\z_o,\z_q)-\D_M^2(\z_o,\z_p) \geq 1+ E [\D_M^2(\x_i,\z_o)]\]
Once the metric is observed, the margin for the expected distances between original data (i.e., as in Eqn.~\ref{eq:ocons}) is also guaranteed. Compared with the original constraints, the margin between latent examples is increased by the factor of $E[\D_M^2(\x_i,\z_o)]$. This term indicates the expected distance between the original data and its corresponding latent example. It means that the tighter a local cluster is, the less a margin should be increased. Furthermore, each class takes a different margin, which depends on the distribution of the original data and makes it more flexible than a global margin. 

With the set of triplets $\{\z_o^t,\z_p^t,\z_q^t\}$, the optimization problem can be written as
\begin{eqnarray*}
\min_{M\in\S_+^{d\times d},\|M\|_F\leq \delta,\z\in{\R^{d\times m}}} \LL(M,\z)=\sum_t \ell(\z_o^t,\z_p^t,\z_q^t;M)
\end{eqnarray*}
where $m\ll n$ is the number of latent examples. We add a constraint for the Frobenius norm of the learned metric to prevent it from overfitting. $\ell(\cdot)$ is the loss function and the hinge loss is applied in this work.
\begin{align*}
&\ell(\z_o^t,\z_p^t,\z_q^t;M)\\
&=[1+E [\D_M^2(\x_i^t,\z_o^t)] -(\D_M^2(\z_o^t,\z_q^t)-\D_M^2(\z_o^t,\z_p^t)) ]_+
\end{align*}
This problem is hard to solve since both the metric and latent examples are the variables to be optimized. Therefore, we propose to solve it in an alternating way and the detailed steps are demonstrated below.

\subsection{Update $\z$ with Upper Bound}
When fixing $M_{k-1}$, the subproblem at the $k$-th iteration becomes
\begin{align}\label{eq:sub1}
&\min_{\z} \LL(M_{k-1},\z)= \sum_t \Big[1+\underbrace{E [\D_{M_{k-1}}^2(\x_i^t,\z_o^t)]}_{a} \nonumber\\
&-\underbrace{(\D_{M_{k-1}}^2(\z_o^t,\z_q^t)-\D_{M_{k-1}}^2(\z_o^t,\z_p^t))}_{b} \Big]_+
\end{align}
The variable $\z$ appears in both the term of margin $a$ and the term of the triplet difference $b$, which makes it hard to optimize directly. Our strategy is to find an appropriate upper bound for the original problem and solve the simple problem instead.

\begin{thm}\label{th:sub1}
The function $\LL(M_{k-1},\z)$ can be upper bounded by the series of functions $\sum_r \F_r(\z)$. For the $r$-th class, we have
\[ \F_r(\z)= c_1E [\D_{M_{k-1}}^2(\x_i,\z_o)] +c_2+c_3\sum_o\D_{M_{k-1}}^2(\z_o,\z_o^{k-1})\]
where $c_1$, $c_2$ and $c_3$ are constants and $\sum_r \F_r(\z^{k-1}) = \LL(M_{k-1},\z^{k-1})$.
\end{thm}
The detailed proof can be found in Section~\ref{sec:analysis}.

After removing the constant terms and rearrange the coefficients, optimizing $\F_r(\z)$ is equivalent to optimizing the following problem
\begin{align}\label{eq:optz}
&\min_{\z\in\R^{d\times m_r},{\bm \mu}:\mu_{i,o}\in\{0,1\},\sum_o \mu_{i,o}=1}\tilde{\F}_r(\z) =\\
& \sum_{i:y(i)=r} \sum_o \mu_{i,o}\D^2_{M_{k-1}}(\x_i,\z_o)+\gamma\sum_o\D_{M_{k-1}}^2(\z_o,\z_o^{k-1})\nonumber
\end{align}
where ${\bm \mu}$ denotes the membership that assigns a latent example for each original example.

Till now, it shows that the original objective $\LL(M_{k-1},\z)$ can be upper bounded by $\sum_r \F_r(\z)$. Minimizing the upper bound is similar to $k$-means but with the distance defined on the metric $M_{k-1}$. So we can solve it by the standard EM algorithm.

When fixing ${\bm \mu}$, latent examples can be updated by the closed-form solution
\begin{eqnarray}\label{eq:optzz}
\forall o,\quad \z_o = \frac{1}{\sum_{i}\mu_{i,o}+\gamma}(\sum_i \mu_{i,o} \x_i+\gamma\z_o^{k-1})
\end{eqnarray}

When fixing $\z$, ${\bm \mu}$ just assigns each original example to its nearest latent example with the distance defined on the metric $M_{k-1}$
\begin{eqnarray}\label{eq:optzu}
\forall i,\quad \mu_{i,o} = \left\{\begin{array}{cc}1&o=\arg\min_oD^2_{M_{k-1}}(\x_i,\z_o)\\0&o.w.\end{array}\right.
\end{eqnarray}

Alg.~\ref{alg:optz} summarizes the method for solving $\tilde{\F}_r(\z)$.
\begin{algorithm}[!h]
   \caption{Algorithm of Updating $\z$}
   \label{alg:optz}
\begin{algorithmic}
   \STATE {\bfseries Input:} data set $\{X,Y\}$, $\z^{k-1}$, $M_{k-1}$, $\gamma$ and $S$
   \STATE Initialize $\z = \z^{k-1}$
   \FOR{$s=1$ {\bfseries to} $S$}
   \STATE Fix $\z$ and obtain the assignment ${\bm \mu}$ as in Eqn.~\ref{eq:optzu}
   \STATE Fix ${\bm \mu}$ and update $\z$ as in Eqn.~\ref{eq:optzz}
   \ENDFOR
   \RETURN $\z^k=\z$
\end{algorithmic}
\end{algorithm}

\subsection{Update $M$ with Upper Bound}
When fixing $\z^k$ at the $k$-th iteration, the subproblem becomes
\begin{align}\label{eq:sub2}
&\min_{M\in\S_+^{d\times d}} \LL(M,\z^k)=\\
&\sum_t [1+\underbrace{E [\D_M^2(\x_i^t,\z_o^t)]}_{a} -\underbrace{(\D_M^2(\z_o^t,\z_q^t)-\D_M^2(\z_o^t,\z_p^t))}_{b} ]_+\nonumber
\end{align}
where $M$ also appears in multiple terms. With the similar procedure, an upper bound can be found to make the optimization simpler.

\begin{thm}\label{th:sub2}
The function $\LL(M,\z^k)$ can be upper bounded by the function $\H(M)$ which is
\begin{align*}
&\H(M) = \frac{\lambda}{2}\|M-M_{k-1}\|_F^2+\sum_t\Big [1+E [\D_{M_{k-1}}^2(\x_i^t,\z_o^t)]\\
&-(\D_M^2(\z_o^t,\z_q^t)-\D_M^2(\z_o^t,\z_p^t)) \Big]_+
\end{align*}
where $\lambda$ is a constant and $\H(M_{k-1}) = \LL(M_{k-1},\z^k)$.
\end{thm}

Minimizing $\H(M)$ is a standard DML problem. Since the number of latent examples $\z^k$ is small, many existing DML methods can handle the problem well. In this work we solve the problem by SGD but sample one epoch active constraints at each stage. The active constraints contain the triplets of $\z^k$ that incur the hinge loss with the distance defined on $M_{k-1}$. This strategy enjoys the efficiency of SGD and the efficacy of learning with the active set. To further improve the efficiency, one projection paradigm is adopted to avoid the expensive PSD projection which costs $\OO(d^3)$. It performs the PSD projection once at the end of the learning algorithm and shows to be effective in many applications~\cite{chechik2010,QianJZL15}. Finally, since the problem is strongly convex, we apply the $\alpha$-suffix averaging strategy, which averages the solutions over the last several iterations, to obtain the optimal convergence rate~\cite{RakhlinSS12}. The complete approach for obtaining $M_k$ is shown in Alg.~\ref{alg:optm}.
\begin{algorithm}[!h]
   \caption{Algorithm of Updating $M$}
   \label{alg:optm}
\begin{algorithmic}
   \STATE {\bfseries Input:} data set $\{X,Y\}$, $\z^{k}$, $M_{k-1}$, $\delta$, $\lambda$ and $S$
   \STATE Initialize $M_0 = M_{k-1}$
   \STATE Sample one epoch active constraints $\mathcal{A}$ according to $\z^k$ and $M_{k-1}$
   \FOR{$s=1$ {\bfseries to} $S$}
   \STATE Randomly sample one constraint from $\mathcal{A}$
   \STATE Compute the stochastic gradient $g = \nabla\H(M)$
   \STATE Update the metric as $M_s' = M_{s-1} - \frac{1}{\lambda s} g$
   \STATE Check the Frobenius norm $M_s = \Pi_{\delta}(M_s')$
   \ENDFOR
   \STATE Project the learned matrix onto the PSD cone \\$M_k =\Pi_{PSD}(\frac{2}{S}\sum_{s=S/2+1}^S M_s) $
   \RETURN $M_k$
\end{algorithmic}
\end{algorithm}

Alg.~\ref{alg:optall} summarizes the proposed margin preserving metric learning framework. Different from the standard alternating method, we only optimize the upper bound for each subproblem. However, the method converges as shown in the following theorem.
\begin{thm}\label{th:sub3}
Let $(\z^{k-1}$, $M_{k-1})$ and $(\z^k$, $M_k)$ denote the results obtained by applying the algorithm in Alg.~\ref{alg:optall} at $(k-1)$-th and $k$-th iterations respectively. Then, we have
\[\LL(\z^k,M_k)\leq \LL(\z^{k-1},M_{k-1})\]
which means the proposed method can converge.
\end{thm}

\begin{algorithm}[!h]
   \caption{\textbf{Ma}rgin \textbf{P}reserving \textbf{M}etric \textbf{L}earning (MaPML)}
   \label{alg:optall}
\begin{algorithmic}
   \STATE {\bfseries Input:} data set $\{X,Y\}$, $\delta$, $m$, $\gamma$, $\lambda$ and $K$
   \STATE Initialize $M_0 = I$
   \FOR{$k=1$ {\bfseries to} $K$}
   \STATE Fix $M_{k-1}$ and obtain latent examples $\z^k$ by Alg.~\ref{alg:optz}
   \STATE Fix $\z^k$ and update the metric $M_k$ by Alg.~\ref{alg:optm}
   \ENDFOR
   \RETURN $M_K$ and $\z^K$
\end{algorithmic}
\end{algorithm}

\paragraph{Computational Complexity} The proposed method consists of two parts: obtaining latent examples and metric learning. For the former one, the cost is linear in the number of latent examples and original examples as $\OO(mn)$. For the latter one, the cost of sampling an active set dominates the learning procedure. Since the number of iterations is fixed, the complexity of sampling becomes $\min\{\OO(Sm),\OO(m^2)\}$. Therefore, the whole algorithm can be linear in the number of latent examples. Note that the efficiency can be further improved with distributed computing since many components of MaPML can be implemented in parallel. For example, when updating $\z$, each class is independent and all subproblems can be solved simultaneously.

\section{Theoretical Analysis}\label{sec:analysis}
\subsection{Proof of Theorem \ref{th:sub1}}
\begin{proof}
First, for the distance of the dissimilar pair in term $b$ of Eqn.~\ref{eq:sub1}, we have
\begin{align*}
&\D^2_M(\z_o,\z_q)= \D_M^2(\z_o^{k-1},\z_q^{k-1})\\
&+\D_M^2(\z_o,\z_o^{k-1})+2(\z_o - \z_o^{k-1})^\top M(\z_o^{k-1}-\z_q^{k-1})\\
&+\D_M^2(\z_q,\z_q^{k-1})-2(\z_q - \z_q^{k-1})^\top M(\z_o^{k-1}-\z_q^{k-1})\\
&-2(\z_o - \z_o^{k-1})^\top M(\z_q - \z_q^{k-1})\\
&\geq \D_M^2(\z_o^{k-1},\z_q^{k-1})-2\D_M(\z_o,\z_o^{k-1})\D_M(\z_o^{k-1},\z_q^{k-1})\\
&-2\D_M(\z_q,\z_q^{k-1})\D_M(\z_o^{k-1},\z_q^{k-1})
\end{align*}
where $\z^{k-1}$ are latent examples from the last iteration. We let $M$ denote $M_{k-1}$ in this proof for simplicity.
The inequality is from that $M$ is a PSD matrix and can be decomposed as $M=LL^\top$. Then it is obtained by applying the  Cauchy-Schwarz inequality.
With the assumptions that $\forall o,\D_M(\z_o,\z_o^{k-1})$ is sufficiently large and $\D_{M}(\z_o^{k-1},\z_q^{k-1})$ is bounded by a constant $\frac{c}{2}$, the inequality can be simplified as
\begin{align}\label{eq:optz1}
&\D^2_M(\z_o,\z_q)\geq \\
&\D_M^2(\z_o^{k-1},\z_q^{k-1})-c\D_M^2(\z_o,\z_o^{k-1})-c\D_M^2(\z_q,\z_q^{k-1})\nonumber
\end{align} 
The assumption is easy to verify since
\[\D_{M}(\z_o^{k-1},\z_q^{k-1})\leq \|\z_o^{k-1}-\z_q^{k-1}\|_2^2\|M_{k-1}\|_2\]
Note that $\|M_{k-1}\|_2\leq \|M_{k-1}\|_F\leq \delta$ and $\z$ is in the convex hull of the original data, and the constant $c$ can be set as $c = 8\delta\max_i\|\x_i\|_2^2$.

With the similar procedure, we have the bound for the distance of the similar pair as
\begin{eqnarray}\label{eq:optz2}
&&\D^2_M(\z_o,\z_p)\leq \D_M^2(\z_o^{k-1},\z_p^{k-1})\\
&&+(c+2)\D_M^2(\z_o,\z_o^{k-1})+(c+2)\D_M^2(\z_p,\z_p^{k-1})\nonumber
\end{eqnarray}

Taking Eqns.~\ref{eq:optz1} and \ref{eq:optz2} back to the original function $\LL(M_{k-1},\z)$ and using the property of the hinge loss, the original one can be upper bounded by
\begin{align*}
&\G(\z)= \sum_t [1+E [\D_M^2(\x_i^t,\z_o^t)] -(\D_M^2(\z_o^{t:k-1},\z_q^{t:k-1})\\
&-\D_M^2(\z_o^{t:k-1},\z_p^{t:k-1}))]_++c_3\sum_o^m \D_M^2(\z_o,\z_o^{k-1})
\end{align*}
where $c_3=\OO(Tc)$ is a constant. 
By investigating the structure of this problem, we find that each class is independent in the optimization problem and the subproblem for the $r$-th class can be written as
\begin{align*}
&\min_{\z\in\R^{d\times m_r}}  \G_r(\z)=\sum_{t:y(\z_o^t)=r} [E [\D_M^2(\x_i,\z_o)]+c_t ]_+\\
&+c_3\sum_{o:y(\z_o)=r}\D_M^2(\z_o,\z_o^{k-1})
\end{align*}
where $m_r$ is the number of latent examples for the $r$-th class and $c_t$ is a constant as
\[c_t = 1-(\D_M^2(\z_o^{t:k-1},\z_q^{t:k-1})-\D_M^2(\z_o^{t:k-1},\z_p^{t:k-1}))\]
Next we try to upper bound the hinge loss in $\G_r(\z)$ with a linear function in the interval of $[c_t,E [\D_M^2(\x_i,\z_o^{k-1})]+c_t]$, where the hinge loss incurred by the optimal solution $\z^k$ is guaranteed to be in it.

\begin{figure}[!ht]
\centering
\includegraphics[width=2.5in]{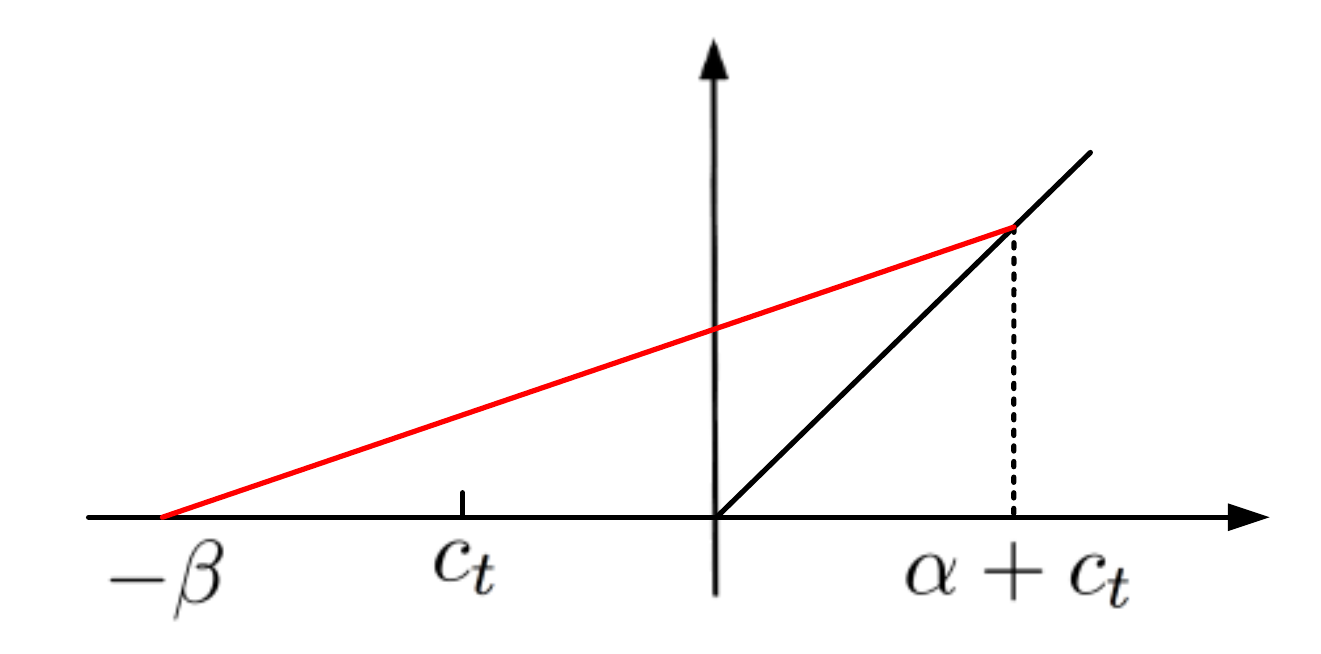}
\caption{\label{fig:hinge}Illustration of bounding the hinge loss. The hinge loss between $[c_t,\alpha+c_t]$ is upper bounded by the linear function denoted by the red line.}
\end{figure}

Let $\alpha = E [\D_M^2(\x_i,\z_o^{k-1})]$, which is the expected distance between the original data of the $r$-th class and the corresponding latent examples from the last iteration, and $\beta$ be a constant sufficiently large as
\[\beta\geq -\min_t c_t\]
Then, for each active hinge loss (i.e., $\alpha+c_t>0$), if
\begin{eqnarray}\label{eq:condi}
E [\D_M^2(\x_i,\z_o)]\leq \alpha
\end{eqnarray}
we have
\begin{eqnarray*}
&&[E [\D_M^2(\x_i,\z_o)]+c_t ]_+\\
&&\leq \frac{\alpha+c_t}{\alpha+c_t+\beta}(E [\D_M^2(\x_i,\z_o)]+c_t+\beta)
\end{eqnarray*}
Fig.~\ref{fig:hinge} illustrates the linear function that can bound the hinge loss and the proof is straightforward. We will show that the condition in Eqn.~\ref{eq:condi} can be satisfied throughout the algorithm later.

With the upper bound of the hinge loss, $\G_r(\z)$ can be bounded by
\begin{align}
\F_r(\z)= c_1E [\D_M^2(\x_i,\z_o)] +c_2+c_3\sum_o\D_M^2(\z_o,\z_o^{k-1})\nonumber
\end{align}
where
\[c_1 = \sum_{t:y(\z_o^{t})=r} \frac{\alpha+c_t}{\alpha_t+c_t+\beta}\mathbbm{I}(\alpha+c_t)\]
and
\[c_2 = \sum_{t:y(\z_o^{t})=r} \frac{\alpha+c_t}{\alpha_t+c_t+\beta}(c_t+\beta)\mathbbm{I}(\alpha+c_t)\]
$\mathbbm{I}(\cdot)$ is an indicator function as $\mathbbm{I}(\nu) = \left\{\begin{array}{cc}1&\nu>0\\0&o.w.\end{array}\right.$

Finally, we check the condition in Eqn.~\ref{eq:condi}. Let $\z^k$ denote latent examples obtained by optimizing $\tilde{\F}(\z)$ with Alg.~\ref{alg:optz}.
Since we use $\z^{k-1}$ as the starting point to optimize $\tilde{\F}_r(\z)$, it is obvious that
\[\tilde{\F}_r(\z^k)\leq \tilde{\F}_r(\z^{k-1})\]
At the same time, we have
\[\sum_o\D_M^2(\z_o^k,\z_o^{k-1})\geq \sum_o\D_M^2(\z_o^{k-1},\z_o^{k-1})=0\]
It is observed that Eqn.~\ref{eq:condi} is satisfied by combining these inequalities.

\end{proof}

\subsection{Proof of Theorem \ref{th:sub2}}
\begin{proof}
For the term $a$ in Eqn.~\ref{eq:sub2}, we have
\begin{align*}
&E [\D_M^2(\x_i,\z_o)] \\
&= E [\D_{M_{k-1}}^2(\x_i,\z_o)+(\x_i-\z_o)^\top(M-M_{k-1})(\x_i-\z_o)]\\
&\leq E [\D_{M_{k-1}}^2(\x_i,\z_o)]+\max_i \|\x_i-\z_o\|_2^2\|M-M_{k-1}\|_F\\
&\leq E [\D_{M_{k-1}}^2(\x_i,\z_o)]+\tilde{c}\|M-M_{k-1}\|_F^2
\end{align*}
where we assume that $\|M-M_{k-1}\|_F$ is sufficiently large and $\tilde{c}$ is a constant which has $\max_i \|\x_i-\z_o\|_2^2\leq \tilde{c}$ and can be set as $\tilde{c} = 4\max_i\|x_i\|_2^2$.

Therefore, the original function $\LL(M,\z^k)$ can be upper bounded by
\begin{align*}
& \H(M) = \frac{\lambda}{2}\|M-M_{k-1}\|_F^2+\sum_t\Big [1+E [\D_{M_{k-1}}^2(\x_i^t,\z_o^t)]\\
&-(\D_M^2(\z_o^t,\z_q^t)-\D_M^2(\z_o^t,\z_p^t)) \Big]_+
\end{align*}
where $\lambda = \OO(T\tilde{c})$.
\end{proof}

\subsection{Proof of Theorem \ref{th:sub3}}
\begin{proof}
When fixing $M_{k-1}$ at the $k$-th iteration, we have
\begin{align*}
&\LL(M_{k-1},\z^k)\leq \sum_r \G_r(\z^k)\leq \sum_r \F_r(\z^k)\\
&\leq \sum_r \F_r(\z^{k-1}) = \LL(M_{k-1},\z^{k-1})
\end{align*}
When fixing $\z^k$, we have
\[\LL(M_k,\z^k)\leq \H(M_k)\leq \H(M_{k-1}) = \LL(M_{k-1},\z^k)\]
Therefore, after each iteration, we have
\[\LL(M_k,\z^k)\leq\LL(M_{k-1},\z^{k-1})\]
Since the value of $\LL(\cdot)$ is bounded, the sequence will converge after a finite number of iterations.
\end{proof}

\section{Experiments}
\label{sec:exp}
We conduct the empirical study on four benchmark image data sets. $3$-nearest neighbor classifier is applied to verify the efficacy of the learned metrics from different methods. The methods in the comparison are summarized as follows.
\begin{itemize}
\item \textbf{Euclid}: $3$-NN with Euclidean distance.
\item \textbf{LMNN}~\cite{weinberger2009}: the state-of-the-art DML method that identifies a set of active triplets with the current metric at each iteration. The active triplets are searched within $3$-nearest neighbors for each example.
\item \textbf{OASIS}~\cite{chechik2010}: an online DML method that receives one random triplet at each iteration. It only updates the metric when the triplet constraint is active.
\item \textbf{HR-SGD}~\cite{QianJY0Z15}: one of the most efficient DML methods with SGD. We adopt the version that randomly samples a mini-batch of triplets at each iteration in the comparison. After sampling, a Bernoulli random variable is generated to decide if updating the current metric or not. With the PSD projection, it guarantees that the learned metric is in the PSD cone at each iteration.
\item \textbf{MaPML$_\tau$}: the proposed method that learns the metric and latent examples simultaneously, where $\tau$ denotes the ratio between the number of latent examples and the number of original ones
\[\tau\% = \frac{m}{n}\]
Different from other methods, $3$-NN is implemented with latent examples as reference points. The method that takes $3$-NN with original data is referred as \textbf{MaPML$_\tau$-O}.
\end{itemize}
The parameters of OASIS, HR-SGD and MaPML are searched in $\{10^{i}:i=-3,\cdots,3\}$. The size of mini-batch in HR-SGD is set to be $10$ as suggested~\cite{QianJY0Z15}. To train the model sufficiently, the number of iterations for LMNN is set to be $10^3$ while the number of randomly sampled triplets is $10^5$ for OASIS and HR-SGD. The number of iterations for MaPML is set as $K=10$ while the number of maximal iterations for solving $M_k$ in the subproblem is set as $S=10^4$, which roughly has the same number of triplets as OASIS and HR-SGD. All experiments are implemented on a server with $96$ GB memory and 2 Intel Xeon E5-2630 CPUs. Average results with standard deviation over $5$ trails are reported.

\subsection{MNIST}
First, we evaluate the performance of different algorithms on MNIST~\cite{lecun1998gradient}. It consists of $60,000$ handwritten digit images for training and $10,000$ images for test. There are 10 classes in the data set, which are corresponding to the digits $0$ - $9$. Each example is a $28\times 28$ grayscale image which leads to the $784$-dimensional features and they are normalized to the range of $[0,1]$. 

\begin{figure}[!ht]
\begin{minipage}{1.5in}
\centering
\includegraphics[height= 1.2 in ]{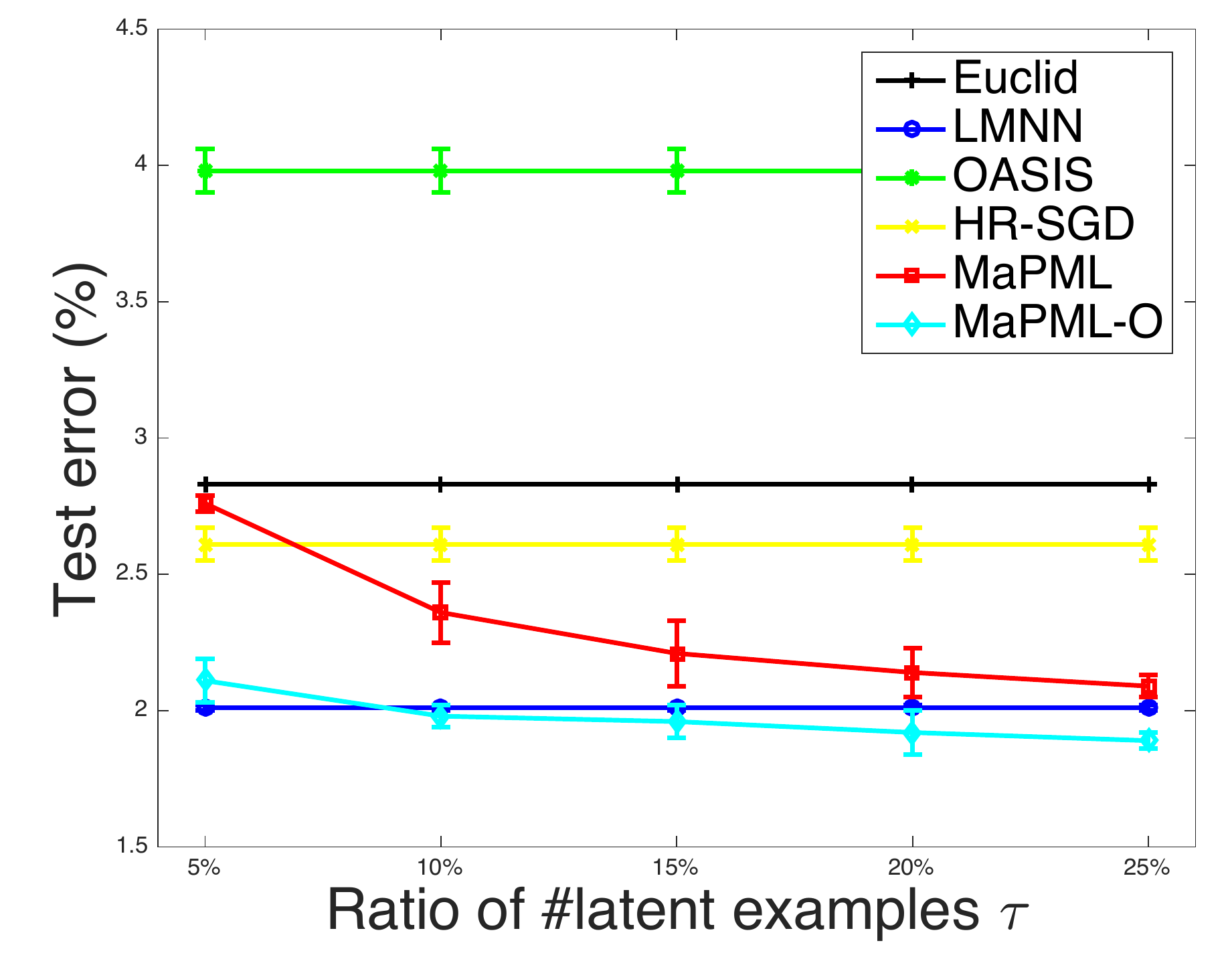}\\
\mbox{\footnotesize (a) Comparison of error rate}
\end{minipage}
\begin{minipage}{1.5in}
\centering
\includegraphics[height= 1.2in ]{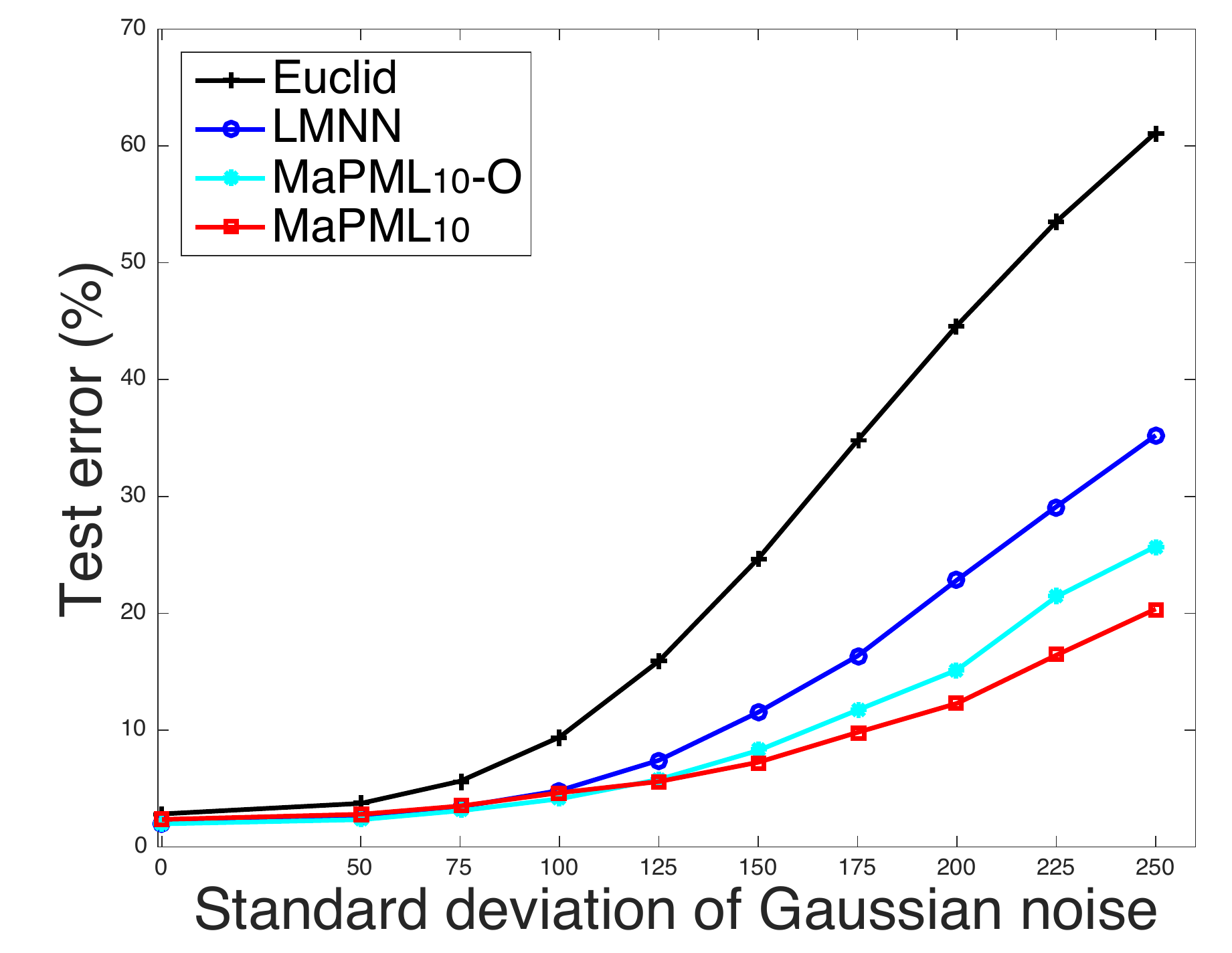}\\
\mbox{\footnotesize (b) Comparison with Gaussian noise}
\end{minipage}
\caption{\label{fig:mnist}Comparisons on MNIST.}
\end{figure}

Fig.~\ref{fig:mnist} (a) compares the performance of different metrics on the test set. For MaPML, we vary the ratio of latent examples from $5\%$ to $25\%$. First of all, It is obvious that the metrics learned with the active set outperform those from random triplets. It confirms that the strategy of sampling triplets randomly can not explore the data set sufficiently due to the extremely large number of triplets. Secondly, the performance of MaPML$_{10}$-O is comparable with LMNN, which shows that the proposed method can learn a good metric with only a small amount of latent examples (i.e., $10\%$). Finally, both MaPML and MaPML-O work well with the metric obtained by MaPML, which verifies that the learned metric can preserve the large margin property for both the original and latent data. Note that when the number of latent examples is small, the performance of $k$-NN with latent examples is slightly worse than that with the whole training set. However, $k$-NN with latent examples can be more robust in real-world applications.

To demonstrate the robustness, we conduct another experiment that randomly introduces the zero mean Gaussian noise (i.e., $\mathcal{N}(0,\sigma^2)$) to each pixel of the original training images. The standard deviation of the Gaussian noise is varied in the range of $[50/255,250/255]$ and $\tau$ is fixed as $10$. Fig.~\ref{fig:mnist} (b) summarizes the results. It shows that MaPML$_{10}$ has the comparable performance as MaPML$_{10}$-O and LMNN when the noise level is low. However, with the increasing of the noise, the performance of LMNN drops dramatically. This can be interpreted by the fact that the metric learned with the original data has been misled by the noisy information. In contrast, the errors made by MaPML and MaPML-O increase mildly and it demonstrates that the learned metric is more robust than the one learned from the original data. MaPML performs best among all methods and it is due to the reason that the uncertainty in latent examples are much less than that in the original ones. It implies that $k$-NN with latent examples is more appropriate for real-world applications with large uncertainty.

\begin{figure}[!ht]
\centering
\begin{minipage}{1.5in}
\centering
\includegraphics[height= 1.2 in ]{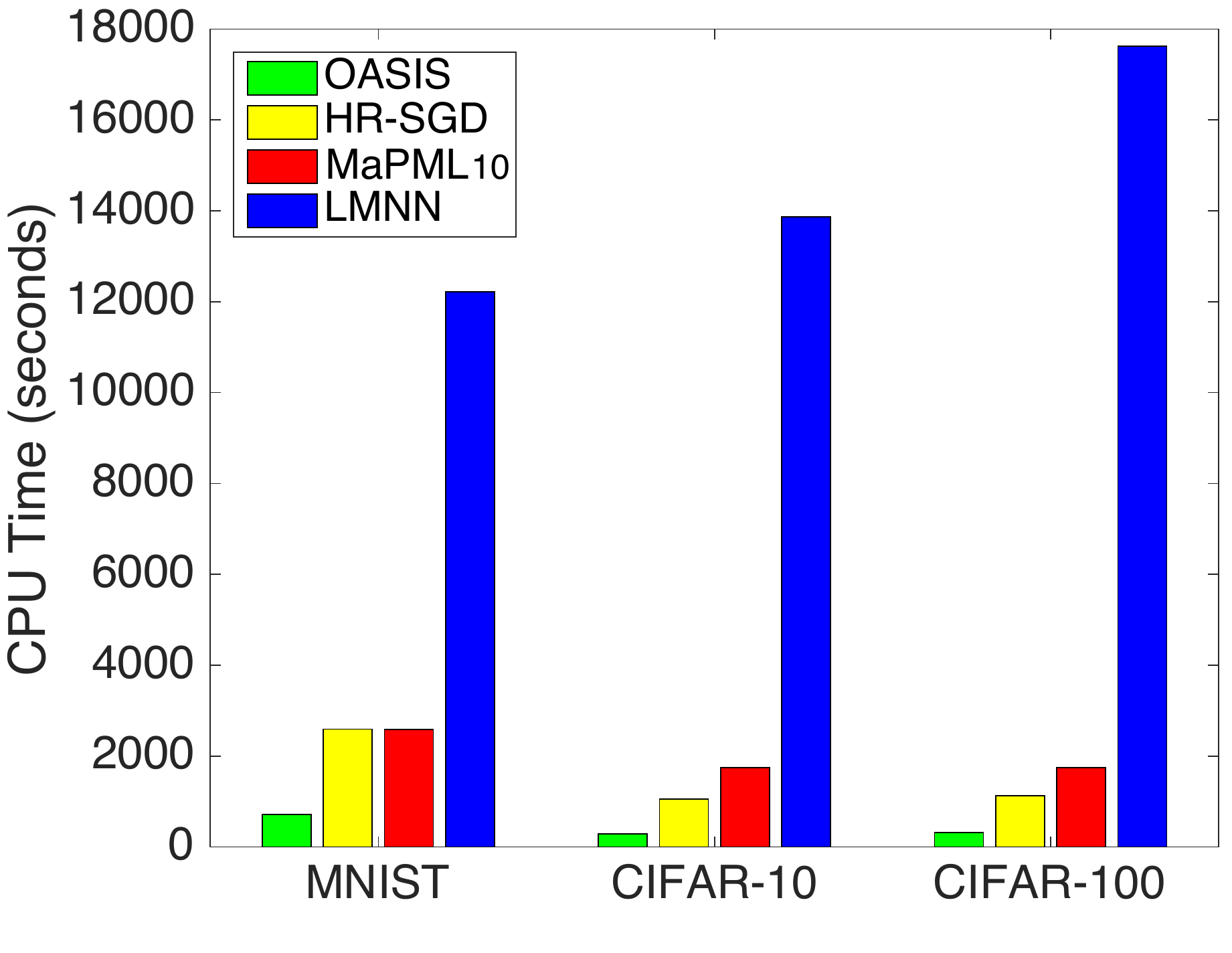}\\
\mbox{\footnotesize (a) CPU time for training }
\end{minipage}
\begin{minipage}{1.5in}
\centering
\includegraphics[height= 1.2in ]{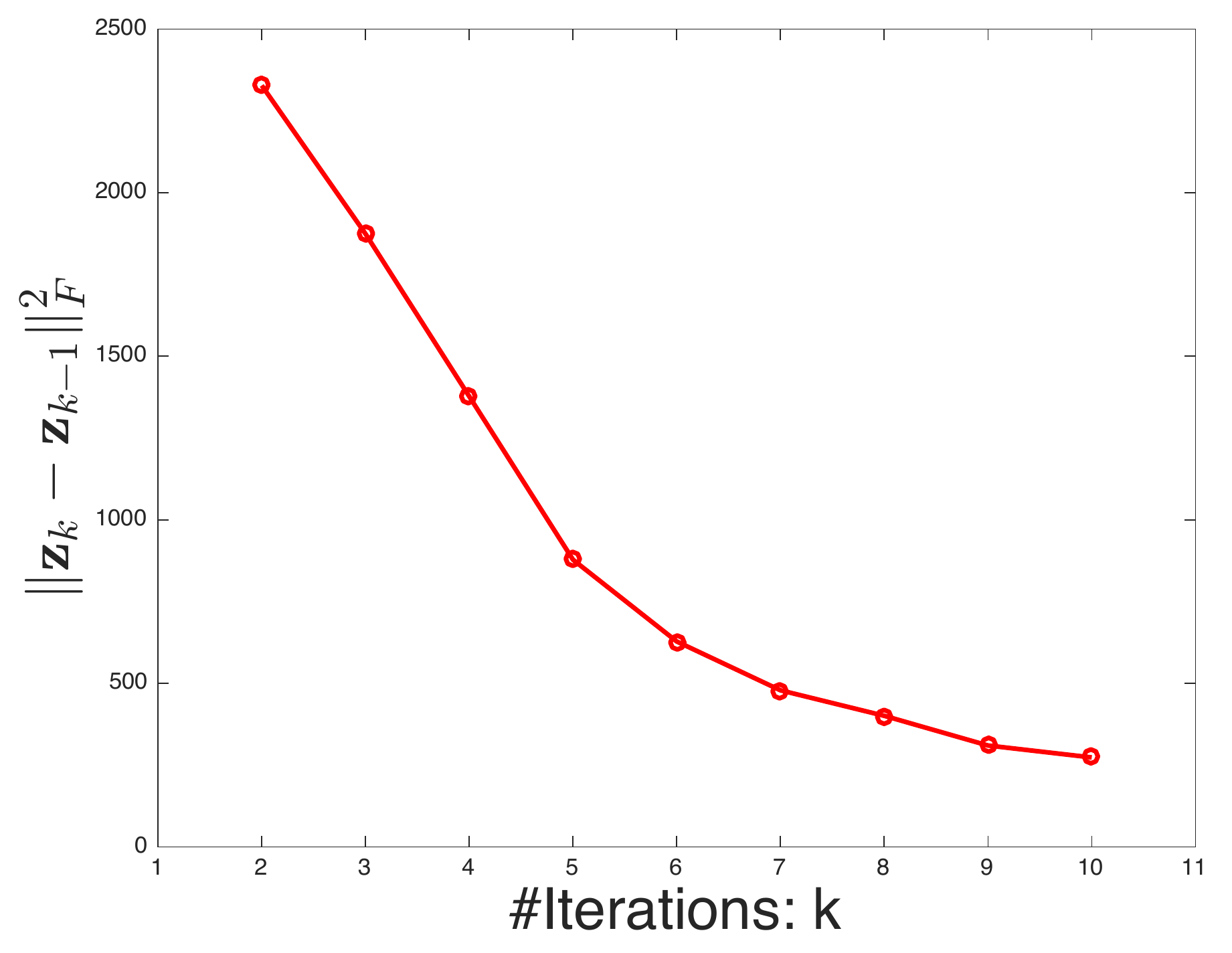}\\
\mbox{\footnotesize (b) Convergence curve of MaPML}
\end{minipage}
\caption{\label{fig:meff}Illustration of the efficiency of the proposed method.}
\end{figure}

Then, we compare the CPU time cost by different algorithms to evaluate the efficiency. The results can be found in Fig.~\ref{fig:meff} (a). First, as expected, all algorithms with SGD are more efficient than LMNN, which has to compute the full gradient from the redefined active set at each iteration. Moreover, the running time of MaPML$_{10}$ is comparable to that of HR-SGD, which shows the efficiency of MaPML with the small set of latent examples. Note that OASIS has the extremely low cost, since it allows the internal metric to be out of the PSD cone. Fig.~\ref{fig:meff} (b) illustrates the convergence curve of MaPML and shows that the proposed method converges fast in practice.

\begin{figure}[!ht]
\centering
\includegraphics[width=3.2in]{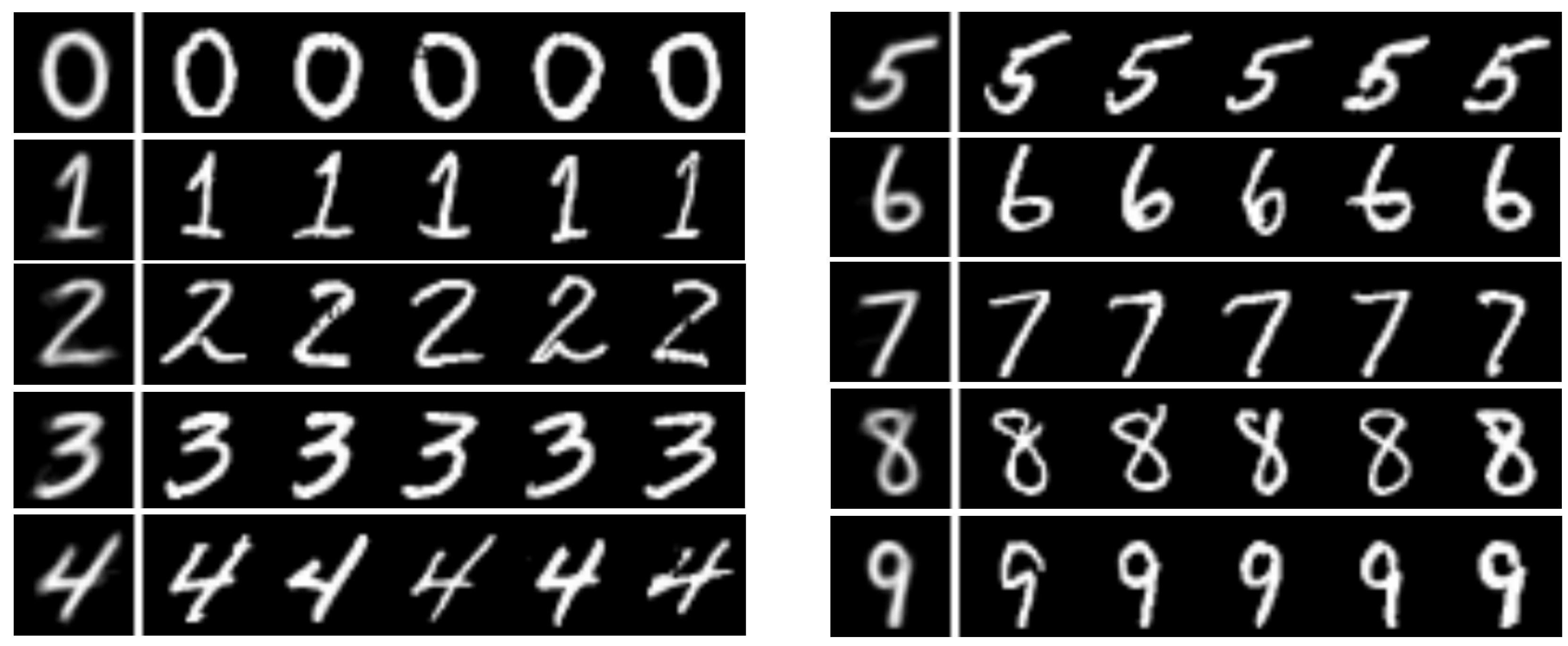}
\caption{\label{fig:mexample}Illustration of the learned latent examples and corresponding original examples from MNIST. The left column indicates latent examples while five original images from each corresponding cluster  are on the right.}
\end{figure}

Finally, since we apply the proposed method to the original pixel features directly, the learned latent examples can be recovered as images. Fig.~\ref{fig:mexample} illustrates the learned latent examples and the corresponding examples in the original training set. It is obvious that the original examples are from latent examples with different distortions as claimed.

\subsection{CIFAR-10 $\&$ CIFAR-100}
CIFAR-10 contains $10$ classes with $50,000$ color images of size $32\times 32$ for training and $10,000$ images for test. CIFAR-100 has the same number of images in training and test but for $100$ classes~\cite{Krizhevsky2009Learning}. Since deep learning algorithms show the overwhelming performance on these data sets, we adopt ResNet18~\cite{He2016Deep} in Caffe~\cite{Jia2014Caffe}, which is pre-trained on ImageNet ILSVRC 2012 data set~\cite{RussakovskyDSKS15}, as the feature extractor and each image is represented by a $512$-dimensional feature vector.

\begin{table}[!ht]
\centering
\caption{Comparison of error rate ($\%$) on CIFAR-10 and CIFAR-100.}\label{ta:c10}
\begin{tabular}{|l||l|l|}\hline
Methods&CIFAR-10&CIFAR-100\\\hline
Euclid&$16.81$&$42.57$ \\\hline
OASIS&$15.22\pm0.18$&$42.46 \pm0.21$\\\hline
HR-SGD&$15.16\pm0.22$&$42.53\pm0.19$\\\hline
LMNN&$13.62\pm 0.12$&$40.05\pm0.13$\\\hline
MaPML$_{10}$-O&$13.59\pm0.14$&$40.49\pm0.15$\\\hline
MaPML$_{10}$&$\mathbf{12.64\pm0.16}$&$\mathbf{34.70\pm0.16}$\\\hline
\end{tabular}
\end{table}


Table \ref{ta:c10} summarizes error rates of methods in the comparison. First, we have the same observation as on MNIST, where the performance of methods adopting active triplets is much better than that of the methods with randomly sampled triplets. Different from MNIST, MaPML$_{10}$ outperforms LMNN on both of the data sets. It is because that images in these data sets describe natural objects which contain much more uncertainty than digits in MNIST. Finally, the performance of MaPML$_{10}$-O is superior over OASIS and HR-SGD, which shows the learned metric can work well with the original data represented by deep features. It confirms that the large margin property is preserved even for the original data.

\subsection{ImageNet}
Finally, we demonstrate that the proposed method can handle the large-scale data set with ImageNet. ImageNet ILSVRC 2012 consists of $1,281,167$ training images and $50,000$ validation data. The same feature extraction procedure as above is applied for each image. Given the large number of training data, we increase the number of triplets for OASIS and HR-SGD to $10^6$. Correspondingly, the number of maximal iterations for solving the subproblem in MaPML is also raised to $10^5$. 

\begin{table}[!ht]
\centering
\caption{Comparison of error rate ($\%$) on ImageNet.}\label{ta:in}
\begin{tabular}{|l||l|}\hline
Methods&Test error ($\%$)\\\hline
Euclid&$35.65$ \\\hline
OASIS&$36.51\pm0.08$\\\hline
HR-SGD&$36.15\pm0.08$\\\hline
MaPML$_{5}$-O&$35.59\pm0.03$\\\hline
MaPML$_{5}$&$\mathbf{33.92\pm0.09}$\\\hline
\end{tabular}
\end{table}

LMNN does not finish the training after 24 hours so the result is not reported for it. In contrast, MaPML obtains the metric within about one hour. The performance of available methods can be found in Table~\ref{ta:in}. Since ResNet18 is trained on ImageNet, the extracted features are optimized for this data set and it is hard to further improve the performance. However, with latent examples, MaPML can further reduce the error rate by $1.7\%$. It indicates that latent examples with low uncertainty are more appropriate for the large-scale data set as the reference points. Note that the small number of reference points will also accelerate the test phase. For example, it costs 0.15s to predict the label of an image with the original set while the cost is only 0.007s if evaluating with latent examples. It makes MaPML with latent examples a potential method for real-time applications. 



\section{Conclusion}
\label{sec:conclusion}
In this work, we propose a framework to learn the distance metric and latent examples simultaneously. By learning from a small set of clean latent examples, MaPML can sample the active triplets efficiently and the learning procedure is robust to the uncertainty in the real-world data. Moreover, MaPML can preserve the large margin property for the original data when learning merely with latent examples. The empirical study confirms the efficacy and efficiency of MaPML. In the future, we plan to evaluate MaPML on different tasks (e.g., information retrieval) and different types of data. Besides, incorporating the proposed strategy to deep metric learning is also an attractive direction. It can accelerate the learning for deep embedding and the resulting latent examples may further improve the performance.
{
\bibliographystyle{ieee}
\bibliography{rdml}

\begin{thebibliography}{10}\itemsep=-1pt

\bibitem{Boyd}
S.~Boyd and L.~Vandenberghe.
\newblock {\em Convex Optimization}.
\newblock Cambridge University Press, New York, NY, USA, 2004.

\bibitem{chechik2010}
G.~Chechik, V.~Sharma, U.~Shalit, and S.~Bengio.
\newblock Large scale online learning of image similarity through ranking.
\newblock {\em JMLR}, 11:1109--1135, 2010.

\bibitem{DavisKJSD07}
J.~V. Davis, B.~Kulis, P.~Jain, S.~Sra, and I.~S. Dhillon.
\newblock Information-theoretic metric learning.
\newblock In {\em ICML}, pages 209--216, 2007.

\bibitem{He2016Deep}
K.~He, X.~Zhang, S.~Ren, and J.~Sun.
\newblock Deep residual learning for image recognition.
\newblock In {\em CVPR}, pages 770--778, 2016.

\bibitem{Jia2014Caffe}
Y.~Jia, E.~Shelhamer, J.~Donahue, S.~Karayev, J.~Long, R.~Girshick,
  S.~Guadarrama, and T.~Darrell.
\newblock Caffe: Convolutional architecture for fast feature embedding.
\newblock In {\em ACM MM}, pages 675--678, 2014.

\bibitem{Krizhevsky2009Learning}
A.~Krizhevsky.
\newblock Learning multiple layers of features from tiny images.
\newblock 2009.

\bibitem{Kulis13}
B.~Kulis.
\newblock Metric learning: A survey.
\newblock {\em Foundations and Trends in Machine Learning}, 5(4):287--364,
  2013.

\bibitem{lecun1998gradient}
Y.~LeCun, L.~Bottou, Y.~Bengio, and P.~Haffner.
\newblock Gradient-based learning applied to document recognition.
\newblock {\em Proceedings of the IEEE}, 86(11):2278--2324, 1998.

\bibitem{Attias17}
Y.~Movshovitz{-}Attias, A.~Toshev, T.~K. Leung, S.~Ioffe, and S.~Singh.
\newblock No fuss distance metric learning using proxies.
\newblock In {\em ICCV}, pages 360--368, 2017.

\bibitem{QianJY0Z15}
Q.~Qian, R.~Jin, J.~Yi, L.~Zhang, and S.~Zhu.
\newblock Efficient distance metric learning by adaptive sampling and
  mini-batch stochastic gradient descent {(SGD)}.
\newblock {\em ML}, 99(3):353--372, 2015.

\bibitem{QianJZL15}
Q.~Qian, R.~Jin, S.~Zhu, and Y.~Lin.
\newblock Fine-grained visual categorization via multi-stage metric learning.
\newblock In {\em CVPR}, pages 3716--3724, 2015.

\bibitem{RakhlinSS12}
A.~Rakhlin, O.~Shamir, and K.~Sridharan.
\newblock Making gradient descent optimal for strongly convex stochastic
  optimization.
\newblock In {\em ICML}, 2012.

\bibitem{RippelPDB15}
O.~Rippel, M.~Paluri, P.~Doll{\'{a}}r, and L.~D. Bourdev.
\newblock Metric learning with adaptive density discrimination.
\newblock In {\em ICLR}, 2016.

\bibitem{RussakovskyDSKS15}
O.~Russakovsky, J.~Deng, H.~Su, J.~Krause, S.~Satheesh, S.~Ma, Z.~Huang,
  A.~Karpathy, A.~Khosla, M.~S. Bernstein, A.~C. Berg, and F.~Li.
\newblock Imagenet large scale visual recognition challenge.
\newblock {\em IJCV}, 115(3):211--252, 2015.

\bibitem{SchroffKP15}
F.~Schroff, D.~Kalenichenko, and J.~Philbin.
\newblock Facenet: {A} unified embedding for face recognition and clustering.
\newblock In {\em CVPR}, pages 815--823, 2015.

\bibitem{SongXJS16}
H.~O. Song, Y.~Xiang, S.~Jegelka, and S.~Savarese.
\newblock Deep metric learning via lifted structured feature embedding.
\newblock In {\em CVPR}, pages 4004--4012, 2016.

\bibitem{weinberger2009}
K.~Q. Weinberger and L.~K. Saul.
\newblock Distance metric learning for large margin nearest neighbor
  classification.
\newblock {\em JMLR}, 10:207--244, 2009.

\bibitem{XingNJR02}
E.~P. Xing, A.~Y. Ng, M.~I. Jordan, and S.~J. Russell.
\newblock Distance metric learning with application to clustering with
  side-information.
\newblock In {\em NIPS}, pages 505--512, 2002.

\bibitem{liu2006}
L.~Yang and R.~Jin.
\newblock Distance metric learning: a comprehensive survery.
\newblock 2006.

\bibitem{YeZSJ17}
H.~Ye, D.~Zhan, X.~Si, and Y.~Jiang.
\newblock Learning mahalanobis distance metric: Considering instance
  disturbance helps.
\newblock In {\em IJCAI}, pages 3315--3321, 2017.

\end{thebibliography}
}

\end{document}